\documentclass[times,referee,twocolumn,final,authoryear]{elsarticle}

\usepackage{ycviu}
\usepackage{framed,multirow}
\usepackage{dblfloatfix}
\usepackage{comment}
\usepackage{natbib}
\setcitestyle{numbers}

\usepackage{amssymb}
\usepackage{latexsym}
\usepackage{amsmath}
\usepackage[justification=centering]{subfig}
\DeclareMathOperator*{\argmin}{arg\,min}

\usepackage{xurl}
\usepackage{xcolor}
\definecolor{newcolor}{rgb}{.8,.349,.1}

\begin{document}

\begin{frontmatter}

\title{Cross-Modal Distillation for RGB-Depth  Person Re-Identification}

\author[1]{Frank M. \snm{Hafner}\corref{cor1}} 
\cortext[cor1]{Corresponding author: 
  Tel.: +4915256191854}
\ead{frank.m.hafner@gmail.com}
\author[2,4]{Amran \snm{Bhuyian}}
\author[3]{Julian F.P. \snm{Kooij}}
\author[2]{Eric \snm{Granger}}

\address[1]{3M Health Information Systems, Hauptstrass 53, Kreuzlingen, Schweiz}
\address[2]{LIVIA, Dept. of Systems Engineering, \'Ecole de technologie sup\'erieure, Montreal, Canada}
\address[3]{Intelligent Vehicles Group, Delft University of Technology, Delft, The Netherlands\\}
\address[4]{Department of CSTE, Noakhali Science \& Technology University, Noakhali-3814, Bangladesh\\}

\received{1 January 2020}
\finalform{10 May 2013}
\accepted{13 May 2013}
\availableonline{15 May 2013}
\communicated{S. Sarkar}

\begin{abstract}
Person re-identification is a key challenge for surveillance across multiple sensors. Prompted by the advent of powerful deep learning models for visual recognition, and inexpensive RGB-D cameras and sensor-rich mobile robotic platforms, e.g. self-driving vehicles, we investigate the relatively unexplored problem of cross-modal re-identification of persons between RGB (color) and depth images. The considerable divergence in data distributions across different sensor modalities introduces additional challenges to the typical difficulties like distinct viewpoints, occlusions, and pose and illumination variation.

While some work has investigated re-identification across RGB and infrared, we take inspiration from successes in transfer learning from RGB to depth in object detection tasks. Our main contribution is a novel method for cross-modal distillation for robust person re-identification, which learns a shared feature representation space of person's appearance in both RGB and depth images. 

In addition, we propose a cross-modal attention mechanism where the gating signal from one modality can dynamically activate the most discriminant CNN filters of the other modality.

The proposed distillation method is compared to conventional and deep learning approaches proposed for other cross-domain re-identification tasks. Results obtained on the public BIWI and RobotPKU datasets indicate that the proposed method can significantly outperform the state-of-the-art approaches by up to 16.1\% in mean Average Precision (mAP), demonstrating the benefit of the distillation paradigm. The experimental results also indicate that using cross-modal attention allows to improve recognition accuracy considerably with respect to the proposed distillation method and relevant state-of-the-art approaches. 

\end{abstract}

\begin{keyword}
\MSC 41A05\sep 41A10\sep 65D05\sep 65D17
\KWD Keyword1\sep Keyword2\sep Keyword3

\end{keyword}

\end{frontmatter}

\section{Introduction}\label{sec:introduction}
\let\thefootnote\relax\footnotetext{Code:   \url{https://github.com/frhf/cross-modal-distillation-reidentification}}

Person re-identification is an important function in many monitoring and surveillance applications, such as multi-camera target tracking, pedestrian tracking in autonomous driving, access control in biometrics, search and retrieval in video surveillance, and forensics~\cite{Gong2014,zheng2016}, and, as such, has gained much attention in recent years. Given the query image of an individual captured using a network of distributed cameras, person re-identification seeks to recognize that same individual over time within a gallery of previously-captured images~\cite{Tan2018}.

\begin{figure}[t]
\centering
\subfloat[Single-modal re-identification]{
\includegraphics[width=.63\linewidth]{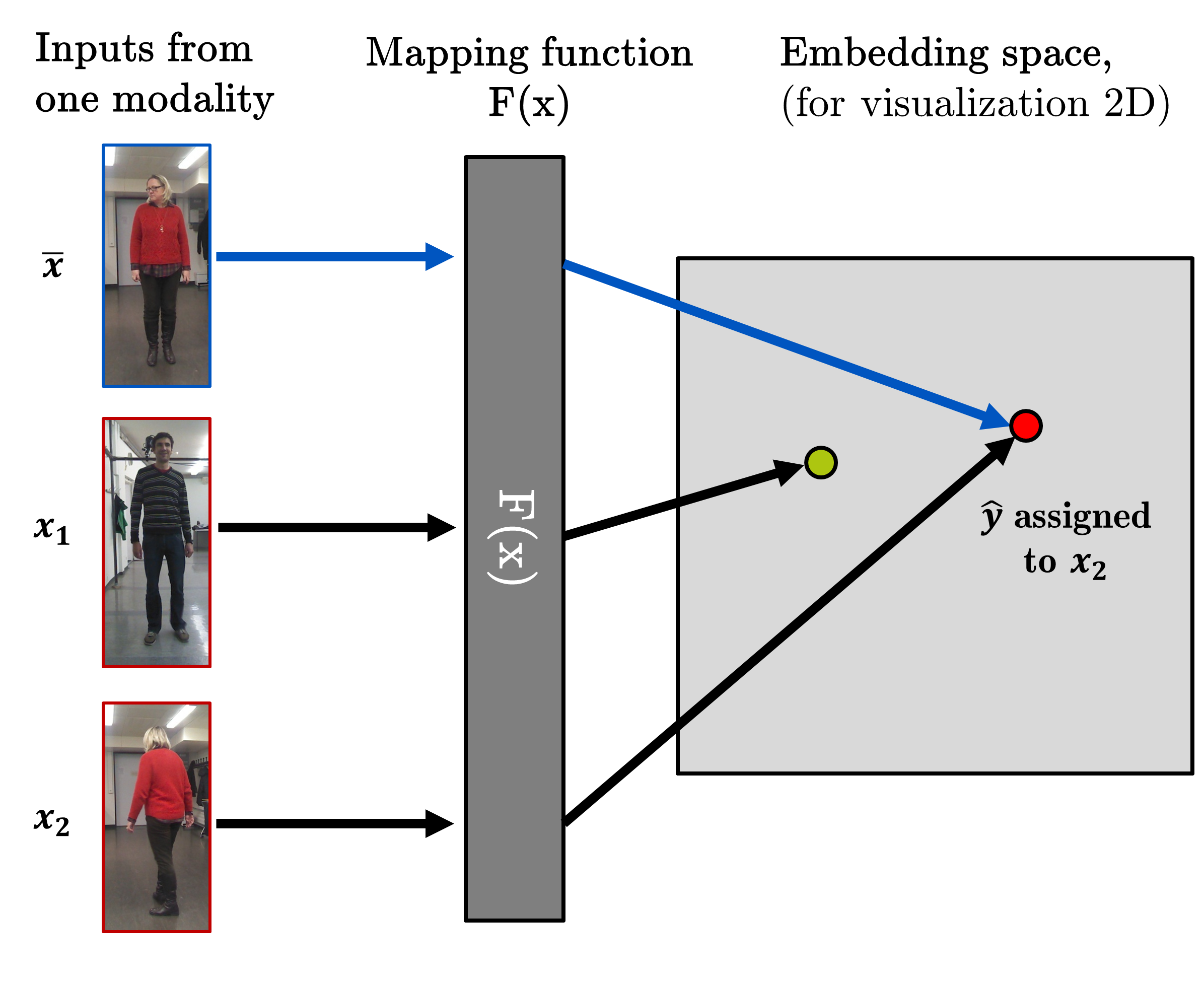}
\label{schemesingmod}
}

\subfloat[Cross-modal re-identification]{
\includegraphics[width=1\linewidth]{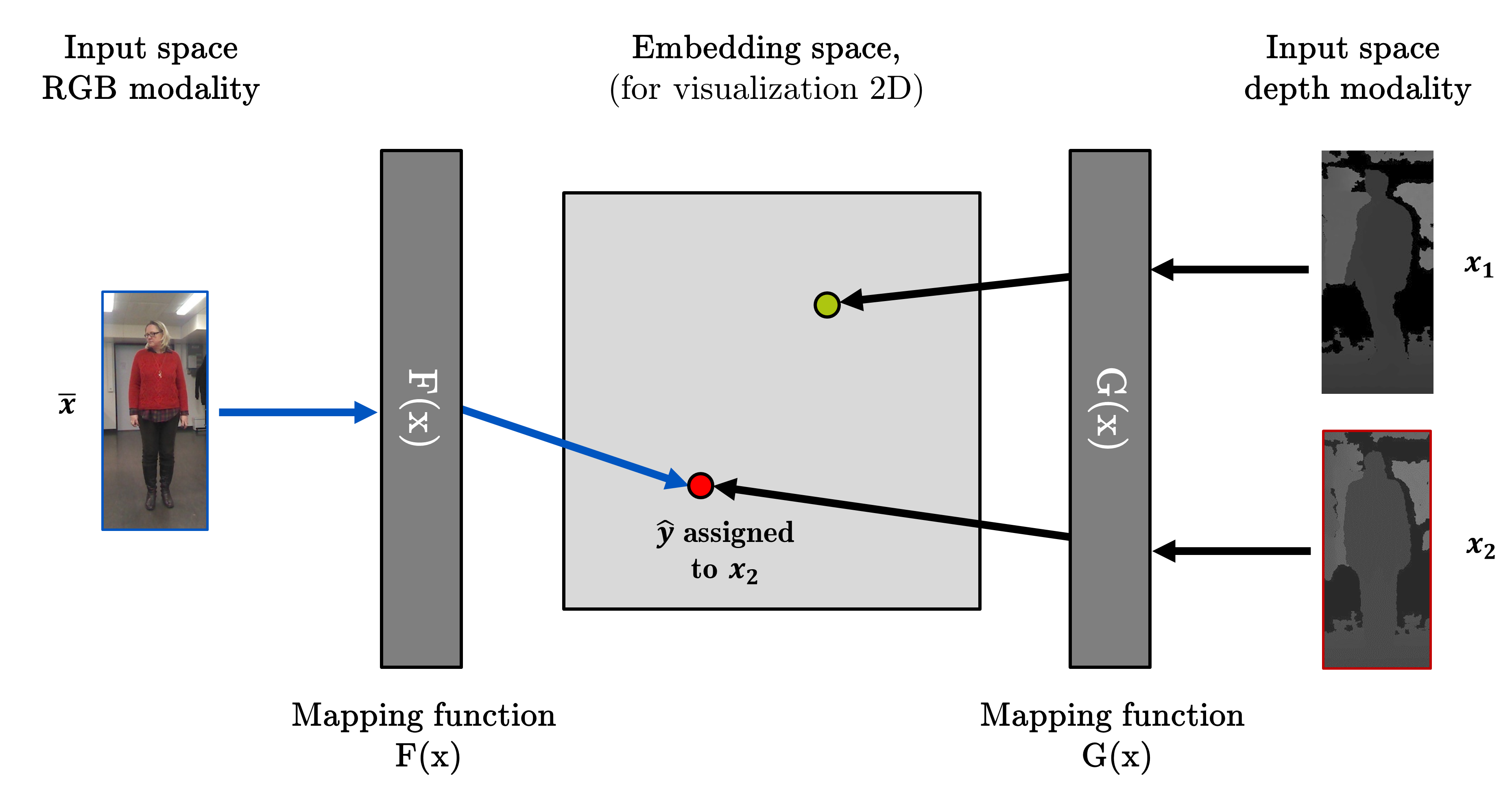}
\label{schemecrossmod}
}
\caption{
\protect\subref{schemesingmod}
Single-modal re-identification embeds input (from the same modality) to a common latent feature space via an embedding function $F(x)$, such that different images from the same individual are close together in the mapping. $\overline{x}$ corresponds to the query image, which needs to be matched with gallery images $x_1$ and $x_2$, where one is from the same class as $\overline{x}$. $\hat{y}$ is associated to the  embedding closest in the latent space to $\overline{x}$.
\protect\subref{schemecrossmod}
Cross-modal re-identification creates a shared embedding for multiple modalities, each with their own mapping function. Here, the embedding functions are defined as $F(x)$ for RGB and $G(x)$ for depth, respectively.}
\label{schemes}
\end{figure}

Re-identification remains a challenging problem in real world applications due to low resolution images, occlusions, miss-alignments, background clutter, motion blur, and variations in pose, scale and illumination.
This paper focuses on the cross-modal variant of this task, which requires matching  a person's appearance across RGB and depth modalities. Figure \ref{schemes} visualizes the difference between the conventional single modality re-identification task (Figure \ref{schemesingmod}) and the cross-modal re-identification task (Figure \ref{schemecrossmod}) addressed in this work.

Although several methods have been proposed for cross-modal re-identification between RGB and infrared images~\cite{wu2017,ye20181,ye20182,dai2018, wang2020}, almost no research addressing RGB and depth images exists and ahead of our work no deep learning based methods have been applied to the task~\cite{zhuo2017,hafner2019}. However, sensing across RGB and depth modalities is important in many real-world scenarios, such as video surveillance systems that must recognize individuals in poorly illuminated environments \cite{Sudhakar2017}. 

Other use cases include robotics and autonomous vehicles, which require tracking pedestrians around their vicinity, where some regions are covered by lidar sensors, and others by RGB cameras \cite{dimi2019,nava2017,nuscenes2020}. Besides these practical applications, research in cross-modal re-identification can also help legal interpretation of depth-based images concerning privacy data protection (e.g. within GDPR)\cite{gdpr2018}. While it is clear that person data from a RGB camera is highly sensible concerning data privacy, it is still unclear how much private information can be extracted from depth images.

In this paper, a new cross-modal distillation training procedure is proposed for robust person re-identification across RGB and depth sensors.
The task is addressed by creating a common embedding of images from both the depth and RGB modalities, as visualized in Figure \ref{schemecrossmod}.
The proposed method exploits a two-step optimization process. In the first step a network is optimized based on data from the first modality, and in the second step the embeddings and weights of this first neural network provide guidance to optimize a second network. The optimization is based on the final embedding layer of the networks to guarantee an embedding in a common feature space for both modalities. The idea behind this approach is to enable a transfer of learned structural representations from the depth modality to the RGB modality, and, therefore, enforce similar feature embeddings for both modalities.

To increase the feature granularity, there is a common trend to use the attention mechanism to address the issue of misalignment in re-identifications. Inspired by the recent success of the gated attention mechanism~\cite{kiran2021,bhuiyan2020,subra2019}, we propose to additionally integrate a cross-modal gated attention mechanism to mitigate the misalignment issue by dynamically selecting the CNN filters. Most of these state-of-the-art approaches use different contextual information to gate the backbone architecture. For instance, \cite{kiran2021} uses optical flow, \cite{subra2019} uses co-segmentation and \cite{bhuiyan2020} uses pose guided contextual information. Unlike~\cite{kiran2021,bhuiyan2020,subra2019}, we introduce the use of cross-modal contextual information, i.e the contextual information from one modality is processed to gate the backbone architecture of another modality. Following the common trend in~\cite{kiran2021,bhuiyan2020,subra2019}, we rely on a simple gated attention mechanism  which allows for multiplicative interaction between the input features from one modality and the attention map from another modality.
This attention is applied into mid-level layer of the respective CNN stream that provide back-propagated gradients corresponding to the amplified local similarities. \\

The contributions of our work are as follows: 
 
~\emph{(i)} A cross-modal distillation training procedure is adopted to transfer an embedding representation from one modality to the other by exploiting the intrinsic relation between depth and RGB. 
\emph{(ii)} We quantitatively and qualitatively show that an ideal deep feature distillation for the task needs to take place from depth to RGB. Hence, the experiments give an understanding of the relationship of the RGB and depth modality. %
\emph{(iii)} We propose to integrate a cross-modal gated attention mechanism into the proposed distillation technique for fine-grained recognition in the embedded space.
\emph{(iv)} An extensive experimental validation is conducted to compare the performance to state-of-the-art methods for cross-modal person re-identification between RGB and depth. On this basis the advantages of the proposed method are shown on multiple RGB-D based benchmark re-identification datasets.

This paper contains a complete discussion of the cross-modal distillation idea for person re-identification between RGB and depth and extends our work in \cite{hafner2019} by comparing to another set of state-of-the-art methods~\cite{wu2017}, conducting an ablation study and a hyperparameter optimization in search of the ideal embedding size. The code to reproduce the results of this paper is published on github.

\section{Related Work}\label{sec:relatedwork}

\begin{figure*}[t]
\centering
\includegraphics[width=1\linewidth]{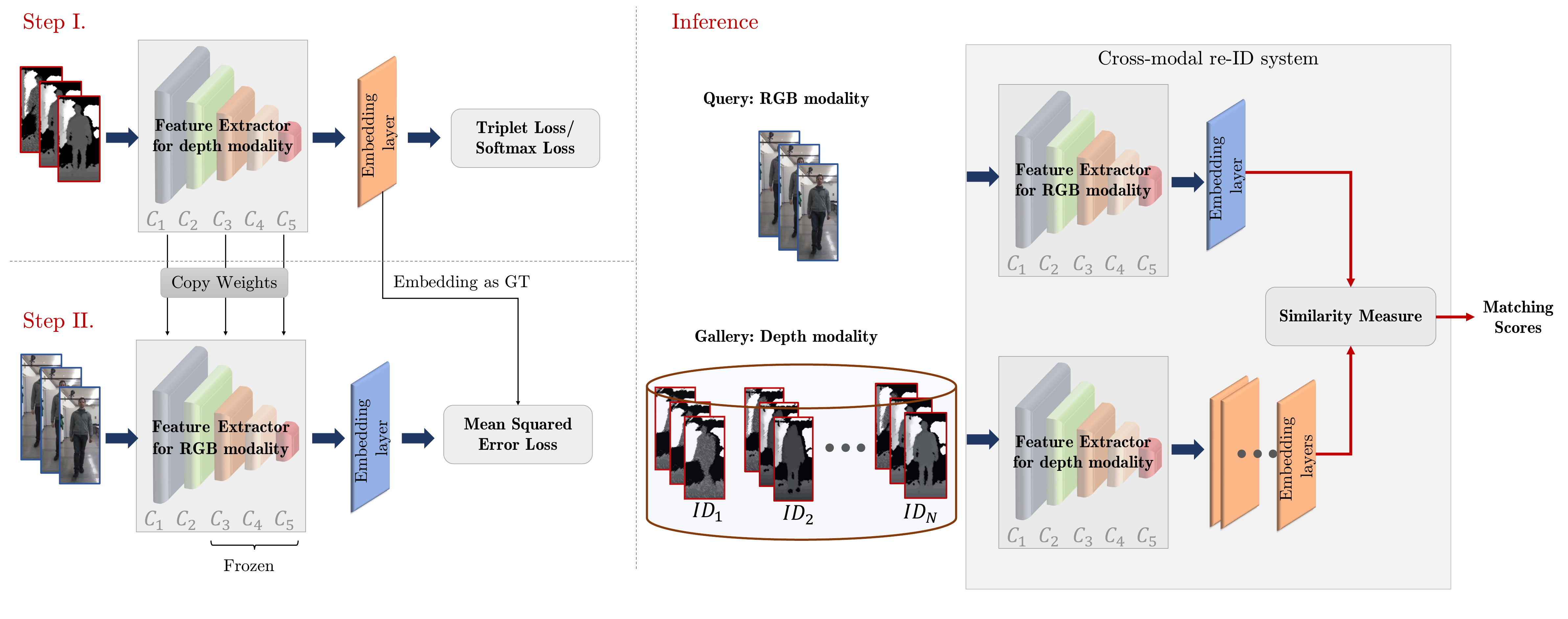}
\caption{Two step training scheme and inference for the proposed training procedure for cross-modal distillation. Step I involves training of a CNN for single-modal re-identification. In step II, the knowledge from the first modality is transferred to the second one. During inference, query and gallery images from different modalities are evaluated to produce feature embeddings and matching scores for cross-modal re-identification. As an example, this Figure shows a transfer from depth to RGB, and inference using RGB as query and depth as gallery. The modalities can be interchanged in both cases.}
\label{schemetransferlearn}
\end{figure*}

The area of person re-identification has received much attention in recent years~\cite{zheng2016}. This section provides a summary of the state-of-the-art conventional, deep learning and cross-modal techniques as they relate to our research.

\textbf{Conventional Methods.} Conventional approaches for person re-identification from a single modality can be categorized into two main groups -- direct methods with hand-crafted descriptors or learned features 
and metric learning based approaches. Direct methods for re-identification are mainly devoted to the search of the most discriminant features to design a powerful descriptor (or signature) for each individual regardless of the scene ~\cite{bhuiyan2014person,liao2015person,panda2017unsupervised}. In contrast, in metric learning methods, a dataset of different labeled individuals is used to jointly learn the features and the metric space to compare them, in order to guarantee a high re-identification rate ~\cite{liao2015person}. 

\textbf{Deep Learning Methods.}
The idea of using a deep learning architecture for person re-identification stems from Siamese CNN with either two or three branches for pairwise verification loss~\cite{Xiao2016} or triplet loss~\cite{Hemans2017,Ergys2018}, respectively. Some of those approaches use their own network architectures, by proposing new layers~\cite{Ahmed2015} or by fusing features from  different body parts with a multi-scale CNN structure~\cite{Cheng2016,Li2017}.
Another stream of works addresses the problem with \textit{transfer learning}~\cite{Geng2016,Xiao2016,Li2018}. Here, the distribution of the training data from the source domain is different from that of the target domain. The most common deep transfer learning strategy for re-identification ~\cite{Geng2016} is to pre-train a base network on a large scale or combination of different datasets as source dataset, and transfer learned representation to the target dataset. A variant of other transfer learning approaches for re-identification~\cite{Xiao2016,Li2018} leverages the idea of joint or multi-task learning considering combination of different re-identification datasets, or auxiliary datasets to minimize the cross-domain discrepancy. 
However, these transfer learning methods depend on the assumption that the tasks are the same and in a single modality and, hence unsuitable when the source and target domains are heterogeneous.

\textbf{Cross-Modal Methods.}
While the progress in re-identifying persons in single modalities was significant, lately the task of cross-modal re-identification receives more attention. Several works~\cite{wu2017,ye20181,ye20182,dai2018,wu20172,zhuo2017, wang2020} investigated the task of cross-modal person re-identification. Recently, several approaches have been proposed for cross-modal person re-identification between RGB and infrared images~\cite{wu2017,ye20181,ye20182,dai2018, wang2020, fan2020,Wang2020,Lu2020}. To embed the RGB and IR modalities in a common feature space, the authors in~\cite{wu2017,ye20181,ye20182} analyze several neural networks architectures: zero padding and one-stream networks~\cite{wu2017}, and two streams network~\cite{wu2017,ye20181,ye20182} with different losses. Additionally, the problem has been addressed in adversarial way in~\cite{dai2018, wang2020}. Interestingly,~\cite{wang2020,Lu2020} identified paired images for training is crucial for the re-identification problem within RGB-infrared and construct those via a GAN. Recently, the problem has been addressed by exploring the potential of both the modality-shared information and the modality-specific characteristics.  Another group of cross-modal re-identification approach~\cite{yan2018,zheng2020,Farooq2020} considers a natural language description of an individual with his/her visual cues. Most of these approaches~\cite{zheng2020,Farooq2020} jointly optimise the two modalities by adapting different matching strategies. 
There are a few works in the literature that consider a \textit{multi-modal} person re-identification scenario~\cite{Enzweiler2011,wu20172} by fusing the RGB and the depth information in order to extract robust discriminative features. 
In~\cite{wu20172}, a depth-shape descriptor called eigen-depth is proposed to extract describing features from the depth domain. \cite{zhuo2017} used the same features to perform cross-modal re-identification between depth and RGB. \cite{uddin2019} combined a body partitioning method and a HOG based feature extraction with a metric learning approach to enable cross-modal person re-identification.

In contrast to the approaches described above for cross-modal re-identification, we propose to employ the cross-modal distillation idea by means of a deep transfer learning technique. The idea of the method is inspired by the work on supervision transfer of Gupta et al.~\cite{gupta2016}. However, supervision transfer~\cite{gupta2016} and our approach aim at different problems with different focuses of method design: supervision transfer solves the problem of limited data availability for object detection problems with a transfer scheme from RGB to depth. Our method is using the distillation paradigm to transfer knowledge from one modality to a second modality to solve the re-identification task across the two modalities. Therefore, contrary to Gupta et al. \cite{gupta2016}, the task has to be solved across modalities in the same feature space and is not considered a pre-training procedure as in \cite{gupta2016}. Additionally, in Gupta et al. the direction of transfer is defined as from RGB to depth. In contrast, in this work the ideal direction of transfer is investigated in detail and a transfer from depth to RGB is shown to be superior for the application. 

\section{Cross-Modal Distillation for Re-Identification}
\label{sec:met}

In this section the cross-modal distillation approach is presented. The approach is used for training of neural networks for cross-modal person re-identification between RGB and depth and is trained with labeled image data from both modalities. During inference, the trained networks then allow to recognize the same person captured using either the RGB or depth sensor. This work extends our previous work \cite{hafner2019} as the first work to apply deep neural networks architectures to solve cross-modal person re-identification between RGB and depth. 

\subsection{Task description}
Consider a query image $\hat{x}$, and a set of gallery images $x_1, \cdots, x_M$ with associated labels $y_1, \cdots, y_M$, such that $y_i$ indicates the individual present in image $x_i$. Each image contains a single individual cropped from a larger input image. 
In single-modal re-identification, both query $\hat{x} \in \chi$ and gallery images $x_i \in \chi$ are from the same input space $\chi$. 
The general approach to person re-identification is to apply a mapping from the input images to an embedded space, where input samples of the same individual are mapped close together, and of different individuals are further apart.
Figure \ref{schemesingmod}
shows how this embedding is used during test time for the standard single-modal case with RGB colour images.
The query image $\hat{x}$ is mapped to the embedded space $F(\hat{x})$,
where the distances to the gallery images $F(x_i)$ are compared.
The identified person $\hat{y}$ for query $\hat{x}$ is then the individual corresponding to the closest embedded gallery image $\hat{i}$, i.e.
\begin{align}
	\hat{y} &= y_{\hat{i}} & \textrm{where}\qquad \hat{i} = \argmin_{i} d(F(\hat{x}), F(x_i)).
    \label{equaone}
\end{align}
where $d$ is the distance metric for the embedding,
typically the Euclidean distance $d(a,b) = \lVert a - b \rVert$.
During training, the learning objective is therefore to estimate a suitable mapping $F(x)$ from available training data.

For cross-modal re-identification an additional challenge is added, namely query $\hat{x} \in \chi_{m1}$ and gallery images $x_i \in \chi_{m2}$ are from separate input spaces $\chi_{m1}$ and $\chi_{m2}$, e.g. $\chi_{m1}$ is the  RGB modality and $\chi_{m2}$ the depth modality.
Figure \ref{schemecrossmod} shows an example with a depth image as query, using RGB gallery images. Since both input spaces have to be mapped to the same latent space, training involves the additional challenge of learning a mapping $G(x)$ for depth images to the feature space shared with RGB mapping $F(x)$. Therefore, Equation \ref{equaone} with RGB query and depth gallery images changes to
\begin{align}
	\hat{y} &= y_{\hat{i}} & \textrm{where}\qquad \hat{i} = \argmin_{i} d(F(\hat{x}), G(x_i)).
    \label{equatwo}
\end{align}

\subsection{Training and inference}
The problem of cross-modal person re-identification is addressed by our novel cross-modal training and inference. 
The training of the networks is divided into two steps to exploit the relationship between depth and RGB and visualized in Figure~\ref{schemetransferlearn}. 
The steps are generic, hence we can either treat RGB as the source and depth as the target domain, or vice versa. To allow for transfer between three-channel RGB and single-channel depth, the inputs are homogenized by duplicating the depth channel into three identical ones.

In step I of a training with cross-modal distillation, a neural network $F$ is trained for sensing in a first modality. 
Afterwards, the network is frozen as $F_{fr}$,
with corresponding weights $W_{F,fr.}$.
In step II, 
a new network $G$ for the second modality is created that maps to the same embedded space as $F$.
We create the new network $G$ using the same architecture as the trained network $F$ from step I.
Similarly to \cite{gupta2016}, the weights of the converged model from step I, $W_{F,fr.}$, are copied to $G$ to serve as initialization.
Additionally, the weights of the mid-level convolutional layer up to the final feature embedding in $G$ are frozen,
As a result, network $G$ can relearn low-level features in the second modality, but retains the high-level embedded space of the first network.
With this approach, we restrict the learning in the second modality and, hence, force the network to learn to extract similar features in the second modality as in the first modality.  
\begin{figure*}[htb!]
\centering
\includegraphics[height=.4\linewidth]{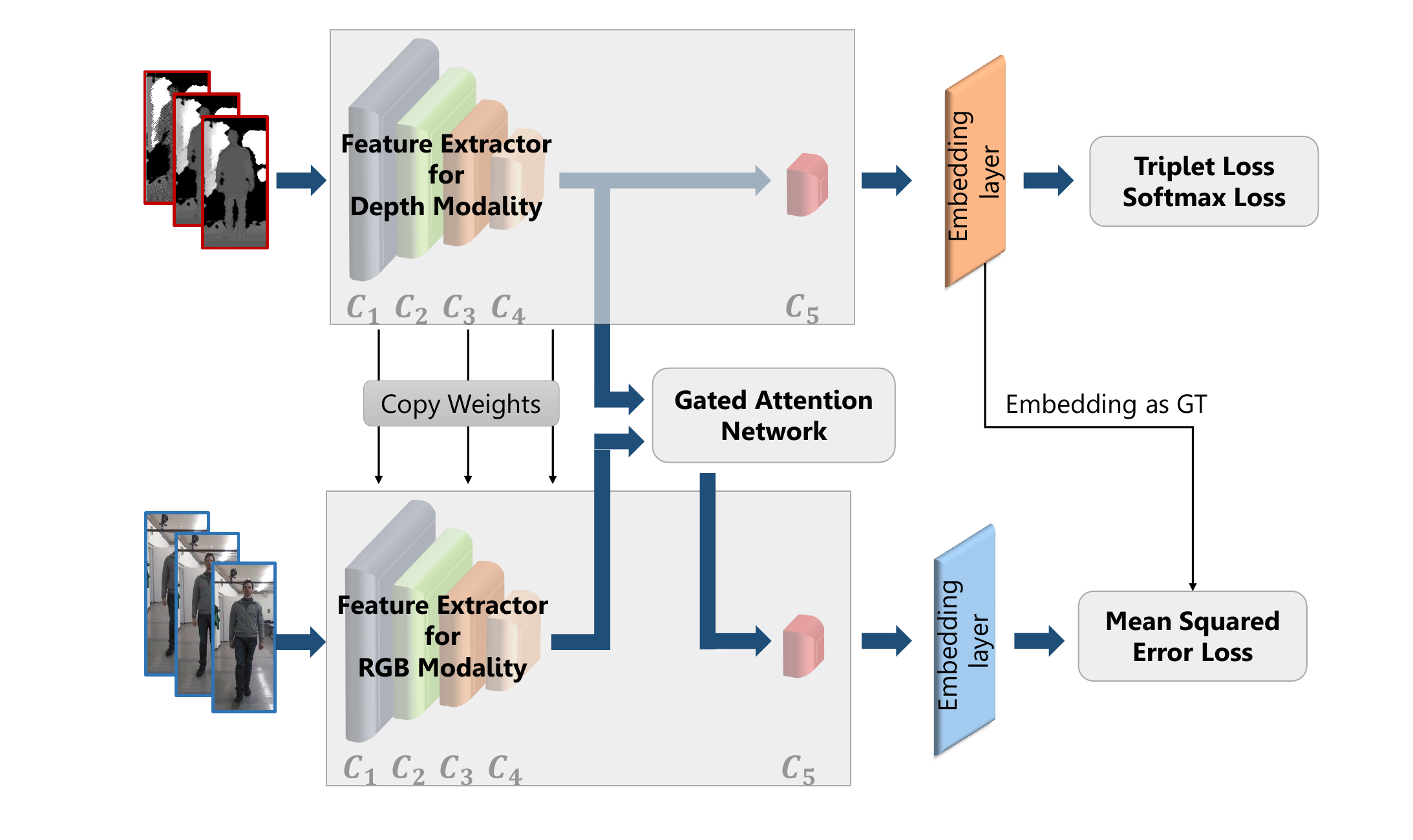}
   \caption{Illustration of the proposed gated attention network.}
\label{fig:attention}
\vspace{-.2cm}
\end{figure*}

For the actual transfer of knowledge we make use of paired images $X_{m1}$ from modality 1 and $X_{m2}$ from modality 2. The aim is to optimize $G$ in such a way that the embeddings of images from the second modality $X_{m2}$ with label $y$ are close to the embeddings of images from the first modality $X_{m1}$ with the same label $y$. This is realized by exploiting image pairs $x_{m1,i}$ and $x_{m2,i}$ from the two modalities, which are considered coupled as they are taken at the exactly same time step.
Hence, the embedding of $x_{m1,i}$ is obtained with a forward propagation through the frozen network $F_{fr.}$ and is taken as the groundtruth for the embedding of $x_{m2,i}$ with the, at this stage, partly optimizable weights of network $G$. Since during inference mode the embeddings will be compared based on Euclidean distance, we aim to minimize this metric between the two embeddings. Hence, we make use of the mean squared error (MSE) loss between the embeddings of paired images $F_{fr.}(x_{m1,i})$ and $G(x_{m2,i})$:

{\small{
\begin{align}
L_{MSE} = \frac{1}{N} \sum^N_{i=1} \lVert F_{fr.}(x_{m1,i})-G(x_{m2,i}) \rVert^2
\label{MSE}
\end{align}
}}
where $N$ is the batch size for training.

During inference, the two resulting networks, $F_{fr.}$ and $G_{fr.}$, are evaluated in the corresponding modalities to provide feature embeddings for input images. Similarity between these representations is measured using Euclidean distance as introduced in Equation \ref{equatwo}. No coupled images are needed for inference.

\subsection{Gated Attention Network}
A gated attention network typically receives a gating signal from another module that provides contextual information~\cite{kiran2021,bhuiyan2020,subra2019}. We here propose to use a gating signal from one modality to act on the backbone of the other modality during training and thereby especially reinforce the features needed for cross-modal matching.
At test time, the input will only contain a single modality, and thus then the attention is computed on that input modality instead, i.e. at test time the network performs self-attention.

In particular, we use depth to gate the backbone of the RGB backbone during training. The task of the gated fusion network is to enable the backbone network to learn informative feature by fusing RGB and depth features within local receptive fields at a fused layer.
The gated attention network, illustrated in Figure~\ref{fig:attention}, takes two inputs: RGB-based appearance features, 	$\textbf{F}^l$ of dimension  $b \times  c_l \times h'\times w'$ from l-th layer of backbone architecture and the depth features, $\textbf{G}^l$ of dimension  $b \times  c_l \times h'\times w'$  from the depth-stream. To get the gated signal, the extracted features from the l-th layer of the depth stream are pooled in the channel dimension to produce features for attention in the spatial dimension. The features are then activated by a function to produce a spatial soft attention map: $\textbf{a}^l = \Psi(\textbf{G}^l )$;
where $\Psi$ is the Sigmoid activation function, and $\textbf{a}^l$ is the gated output of the activation function of size $b \times  1 \times h'\times w'$.
We use a simple and effective scheme to align the RGB-based appearance feature using that gated signal. First, we compute $\textbf{F}^l_a$, the Hadamard product (element-wise multiplication) between RGB-based appearance feature $\textbf{F}^l$ and gated attention $\textbf{a}^l$. $\textbf{F}^l_a$ is then added to the original $\textbf{F}^l$ to obtain the final aligned attention map $\textbf{F}^l_g$ for the remaining layers in the backbone network of the  RGB-stream:
\begin{equation}\label{gate_features}
	\begin{aligned}
	\textbf{F}^l_a  = \textbf{F}^l \bigotimes \textbf{a}_l \ \  \ \  \textbf{F}^l_g= \textbf{F}^l_a + \textbf{F}^l
	\end{aligned}
\end{equation}
\noindent where $\bigotimes$ denotes element-wise Hadamard product.

At test time the attention is instead computed on the feature map of the now uni-modal input, $\textbf{a}^l = \Psi(\textbf{F}^l )$,
since the training procedure has ensured that the feature map of the input modality closely matches the map that the other modality would have provided.


\section{Experimental Methodology}
\label{sec:experimentmet}
We here present the implementation of the cross-modal distillation approach, the two RGB-D person re-identification datasets, and the evaluation protocol and metrics that we will use to validate our proposed approach.
While these are public datasets, they were not originally designed for cross-modal person re-identification, hence we first present their relevant properties in this context.

\subsection{Implementation}
\label{sec:singmodtech}

For this work a selection of successful feature extraction networks and loss functions will be deployed. 
For the validation of the concepts in this work residual neural networks with 50 layers (ResNet50)~\cite{He2016} which are pre-trained on ImageNet was chosen as a good compromise between depth and accuracy. The ResNet architecture was shown to be effective for several person re-identification applications~\cite{Ergys2018,Zheng2017wild}.
Furthermore, we consider two possible loss functions, triplet loss and softmax loss, which both have been successfully applied in single-modal person re-identification~\cite{zheng2016, Hemans2017}.
We will now shortly discuss both losses in more detail.

Using the \textit{triplet loss} results in a metric learning approach which directly optimizes an embedding layer in Euclidean space.
During training, this loss compares the relative distances of three training samples,
namely an anchor image $x_a$, a positive image sample $x_p$ from the same individual as $x_a$, and a negative sample $x_n$ from a different individual.
Given an anchor image $x_a$, this loss assures that the embedding of an image taken from the same class $x_p$ is closer to the anchor's embedding than that of a negative image belonging to another class $y_n$ by at least a margin $m$ in distance metric $d$.
In the following, $F(x)$ denotes the mapping to the embedded space, which is in our case a deep neural network.
The triplet loss is therefore defined as: 

{\small{
\begin{align}
 L_{tri}= \sum_{i=1}^T \left[ d(F(x_{a(i)}), F(x_{p(i)}))-d(F(x_{a(i)}), F(x_{n(i)}))+m \right].
\label{triplet}
\end{align}}}
where indices $a(i)$, $p(i)$ and $n(i)$ stand for anchor, positive and negative,
of the $i$-th triplet, and $T$ for the number of triplets used per batch.

For the \textit{softmax loss}, 
the embedding is learned indirectly by first treating re-identification on the training set as a classification problem,
where all individuals in the training set are considered a different class.
Afterwards, the layer of the neural network prior to the softmax loss is used as the embedding. This enables that the network can be applied on test data, which can contain new individuals not present in the training data.
 Therefore, the softmax loss to optimize the embedding can be expressed as:

{\small{
\begin{align}
L_{soft}= -\frac{1}{N}\sum^{N}_{i=1} \log \left( \frac{e^{W_{(y_i)} F(x_i) + b}}{\sum^C_{j=1}e^{ W_{(j)} F(x_i) + b}}\right),
\label{softmax}
\end{align}}}
where $N$ is the batch size, $W_{(j)}$ are the weights leading to the $j$-th node of the ultimate softmax layer of the network, $b$ is a bias and the number of classes is defined as $C$.
For triplet loss an embedding size of 128 and a training batch of 64 with 16 instances \'a 4 images was used for the experiments in this work and batch hard mining was chosen for triplet choice. These parameters were proposed by \cite{Hemans2017}. To enable a fair comparison also for softmax loss an embedding size of 128 was chosen. The neural networks trained with softmax were optimized with stochastic gradient descent with Nesterov momentum. Those trained with triplet loss were optimized with the ADAM optimizer. The margin for triplet loss (see Equation \ref{triplet}) was set to 0.5.
The networks which trained in a single-modality are optimized using an early-stopping criterium based on the mean Average Precision (mAP) on the validation set.
For step II of the cross-modal distillation approach as explained in chapter \ref{sec:met} the early-stopping criteria for network training is the loss in validation data.

\subsection{Datasets}
The considered datasets are BIWI RGBD-ID \cite{munaro2014} and RobotPKU \cite{liu2017}. These datasets were selected because they provide high-resolution depth and RGB images, a decent number of person instances and a large amount of images per person instance in different poses. Both datasets were recorded with a Microsoft Kinect camera. The TVPR datasets \cite{TVPR,TVPR2} which contain an increased number of person instances were neglected as we argue that bird's eye view data does not contain enough shape information for cross-modal person re-identification.  

The BIWI RGBD-ID dataset targets long-term people re-identification from RGB-D cameras \cite{munaro2014}.  
As in \cite{zhuo2017} same person with different clothing is considered as a separate instance. Overall, it is comprised of 78 instances with 22,038 images in depth and RGB. Exemplary images from BIWI are shown in Figure \ref{resultsbiwi}  
 RGB and depth images are provided coupled with no visible difference in capturing time.

The RobotPKU dataset consists of 90 persons with 16,512 images in total \cite{liu2017}. 
The images are provided in a coupled manner. Nevertheless, through visual inspection it is apparent that there is a slight time difference in the order of a fraction of a second between the images captured in depth and RGB. For training and inference, images of both datasets in both modalities are resized to $256 \times 128$.
Compared to the BIWI dataset, the depth images in the RobotPKU dataset are more noisy and often body parts, like heads and arms, are lost in the images. Also the movements of the probands are more dynamic and heterogeneous.

\subsection{Evaluation Protocol}
For the performance evaluation with the BIWI dataset, the subset partitions for training, validation and testing were adopted from \cite{zhuo2017}, which means 32 individuals were used for training, 8 instances for validation and 38 individuals for testing. For the RobotPKU dataset, the division will be videos from 40 individuals for training, 10 for validation, and 40 for testing. This follows the division of \cite{liu2017}. The exact split (label of individuals used to form subsets) is provided in appendix A.
For quantitative evaluation, the average rank 1, 5 and 10 accuracy performance measure is reported along with the mean average precision (mAP). To report the rank accuracy, a single-gallery shot setting is used, where a random selection of the gallery (G) images is repeated 10 times. Hereby, the exactly same corresponding image or paired image in the parallel modality is excluded for the random sampling. For calculating the mAP a maximum of 20 images per person in the scenario are randomly selected.

To obtain statistically reliable results we introduce a 3-fold cross-validation process, where a different validation subset is randomly extracted from within the training subset.

\section{Results and Discussion}
\label{sec:experiments}
An extensive series of experiments has been considered to validate the proposed networks trained with cross-modal distillation. In this section, the results for optimization with the single modalities (i.e., step I. in Fig~\ref{schemetransferlearn}) are first shown to establish a baseline for the individual modalities (section \ref{sec:singmod}). Hence, we first investigate how different choices for losses affect the performance on single-modal re-identification, and compare the relative difficulty of the modalities and dataset. Then, the distillation step (step II.) of the proposed method is performed and evaluated (section \ref{sec:crossdist}). Here, the ideal direction of transfer is investigated. In section \ref{sec:embsize} an analysis of the ideal size of the embedding layer will be performed. 
Finally, the state-of-the-art of the cross-modal person re-identification task between RGB and depth is defined (section \ref{sec:compex}) and a qualitative explanation for the superiority of the method will be provided (section \ref{sec:complex}) 
%

\subsection{Single-Modal Re-identification Performance}
\label{sec:singmod}
For performance evaluation with individual modalities (RGB and depth separately), the single-modal case, results have been obtained on BIWI and RobotPKU datasets. The representative feature extractor Resnet50 has been optimized with triplet loss, Equation~\eqref{triplet}, and softmax loss, Equation~\eqref{softmax}.

\begin{table}[t!]
\caption{Average test set accuracy of the baseline single-modal re-identification (Step I) for different modalities on BIWI dataset.}
\centering
\resizebox{0.5\textwidth}{!}
{
\begin{tabular}{ll||lll|l|}
\cline{4-6}
\hline
\multicolumn{1}{|l|}{\textbf{Modality}}                  &  \textbf{Loss}  & \multicolumn{1}{l|}{\textbf{rank-1 (\%)}} & \multicolumn{1}{l|}{\textbf{rank-5 (\%)}} & \multicolumn{1}{l|}{\textbf{rank-10 (\%)}} & \textbf{mAP (\%)} \\ \hline \hline
\multicolumn{1}{|l|}{{\textbf{RGB}}}      &   Triplet    & 92.1 $\pm$ 1.9  & 99.7 $\pm$ 0.2  & 99.9 $\pm$ 0.1  & 93.4 $\pm$ 1.5           \\ 
\multicolumn{1}{|l|}{}                                                                                           & Softmax     &  \textbf{94.8 $\pm$ 0.7} & \textbf{99.8 $\pm$ 0.2} & \textbf{99.9 $\pm$ 0.0} & \textbf{95.7 $\pm$ 0.6}            \\ \cline{1-3}
\hline
\hline
\multicolumn{1}{|l|}{{\textbf{Depth}}}                                                       & Triplet   & 54.2 $\pm$ 1.8  & \textbf{91.5 $\pm$ 0.6}  & \textbf{99.2 $\pm$ 0.2}  & 55.3 $\pm$ 1.7           \\  
\multicolumn{1}{|l|}{}                                                                                    & Softmax  & \textbf{59.8 $\pm$ 0.7} & 90.5 $\pm$ 0.8 & 97.8 $\pm$ 0.2 & \textbf{61.4 $\pm$ 0.5}                \\ \cline{1-3}
\hline
\end{tabular}
}
\label{biwitab}
\end{table}

\begin{table}[t!]
\caption{Average test set accuracy of the baseline single-modal re-identification (Step I) for different modalities on RobotPKU dataset.}
\centering
\resizebox{0.5\textwidth}{!}
{
\begin{tabular}{ll||lll|l|}
\cline{4-6}
\hline
\multicolumn{1}{|l|}{\textbf{Modality}}                  & \textbf{Loss}  & \multicolumn{1}{l|}{\textbf{rank-1 (\%)}} & \multicolumn{1}{l|}{\textbf{rank-5 (\%)}} & \multicolumn{1}{l|}{\textbf{rank-10 (\%)}} & \textbf{mAP (\%)} \\ \hline \hline
\multicolumn{1}{|l|}{{\textbf{RGB}}}                                                                 & Triplet    & \textbf{89.0 $\pm$ 3.9}  & \textbf{99.2 $\pm$ 0.3}  & \textbf{99.5 $\pm$ 0.1}  & \textbf{90.6 $\pm$ 3.4}           \\
\multicolumn{1}{|l|}{}                                                                                            & Softmax       &  84.5 $\pm$ 0.2 & 97.9 $\pm$ 0.4 & 99.1 $\pm$ 0.2 & 87.1 $\pm$ 0.2            \\ \cline{1-3}
\hline
\hline
\multicolumn{1}{|l|}{{\textbf{Depth}}}                                                                       & Triplet   & n/a  & n/a  & n/a  & n/a           \\
\multicolumn{1}{|l|}{}                                                                                             & Softmax  & \textbf{44.5 $\pm$ 1.0} & \textbf{75.8 $\pm$ 1.3} & \textbf{87.6 $\pm$ 0.9} & \textbf{44.5 $\pm$ 1.0}               \\ \cline{1-3}
\hline
\end{tabular}
}

\label{pkulabel}
\end{table}

Table~\ref{biwitab} shows the average accuracy of the networks for single-modal re-identification for single (RGB and depth) modalities on BIWI data and Table ~\ref{pkulabel} the same results for RobotPKU data. Results show that the networks optimized using RGB modality alone can reach a high level of accuracy (95.7\% for BIWI with softmax loss and 90.6\% for RobotPKU with triplet loss). As expected, the overall accuracy for the networks optimized using depth modality alone is much lower compared to the accuracy achieved for the same task with RGB (mAP of 61.4\% for BIWI with triplet loss and 44.5\% for RobotPKU with softmax loss).  Note that on RobotPKU, training with triplet loss did not converge to produce meaningful embedding layers, which we believe is due to the higher level of noise and diversity in that dataset. Overall, RGB has the stronger visual cues for re-identification as expected, but descriptive features can also be extracted from depth. Even though the performance is comparably lower for sensing in pure depth, the achieved performance serves as a baseline with which practical use cases can be approached.

\subsection{Performance for Cross-Modal Distillation}
\label{sec:crossdist}
In this section the experiments with the cross-modal distillation method as presented in section~\ref{sec:met} will be introduced. As baseline or step I of the method the results from section ~\ref{sec:singmod} will be considered.
In this section experiments are presented to gain insight on step II
(distillation), and, in particular, on the advantages of transferring knowledge based on the depth or RGB modality.

\begin{figure}[t!]
\centering
\includegraphics[width=0.95\linewidth]{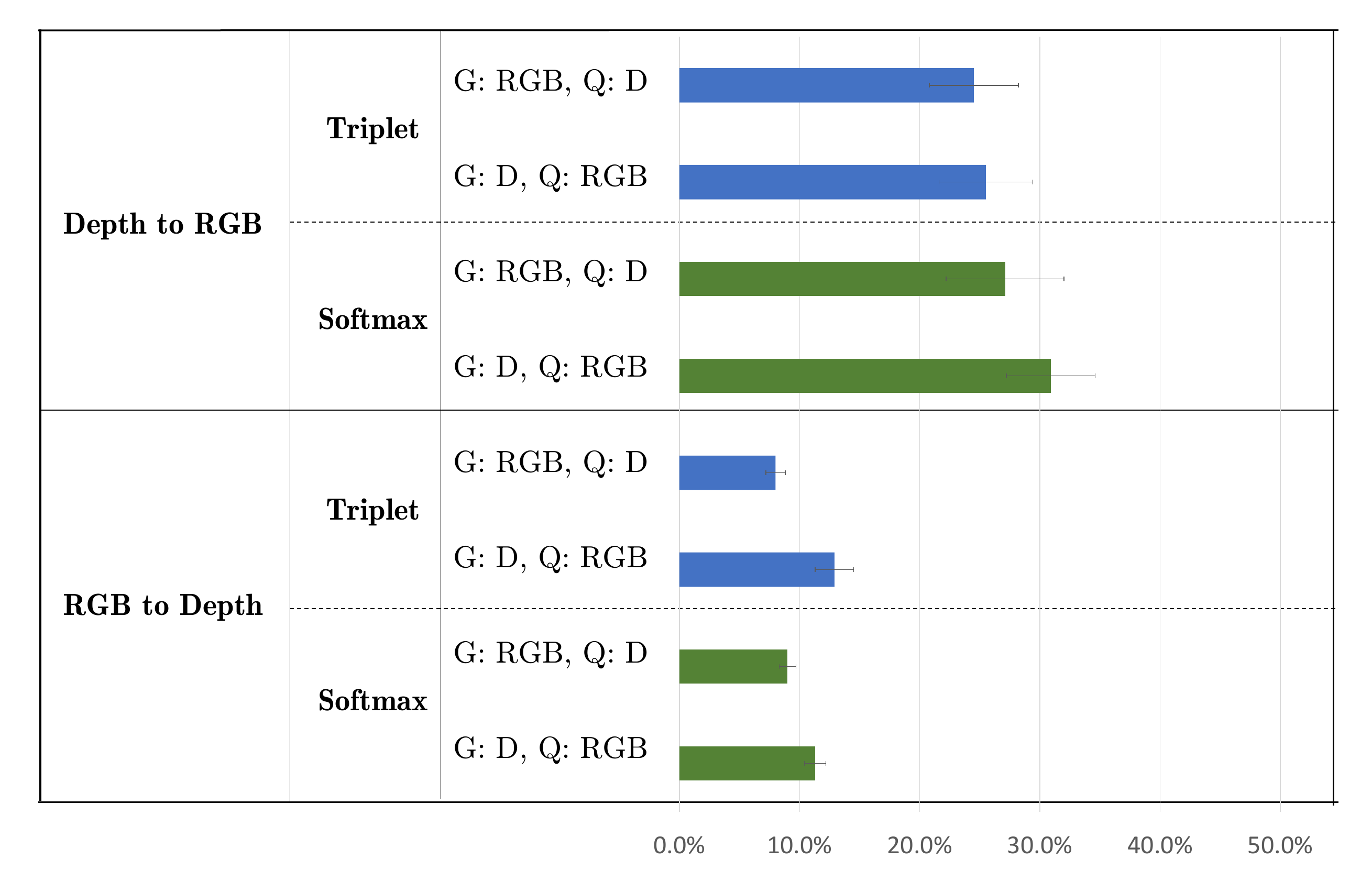}
\caption{Average mAP accuracy of various networks trained with cross-modal distillation on the BIWI dataset. For all combinations we report varying query (Q) and gallery (G) modalities. The first column indicates the direction of the transfer for the cross-modal distillation. The different colors indicate results with triplet (blue) and softmax (green) loss functions.}
\label{biwi_transresults}
\end{figure}

\begin{figure}[t!]
\centering
\includegraphics[width=0.95\linewidth]{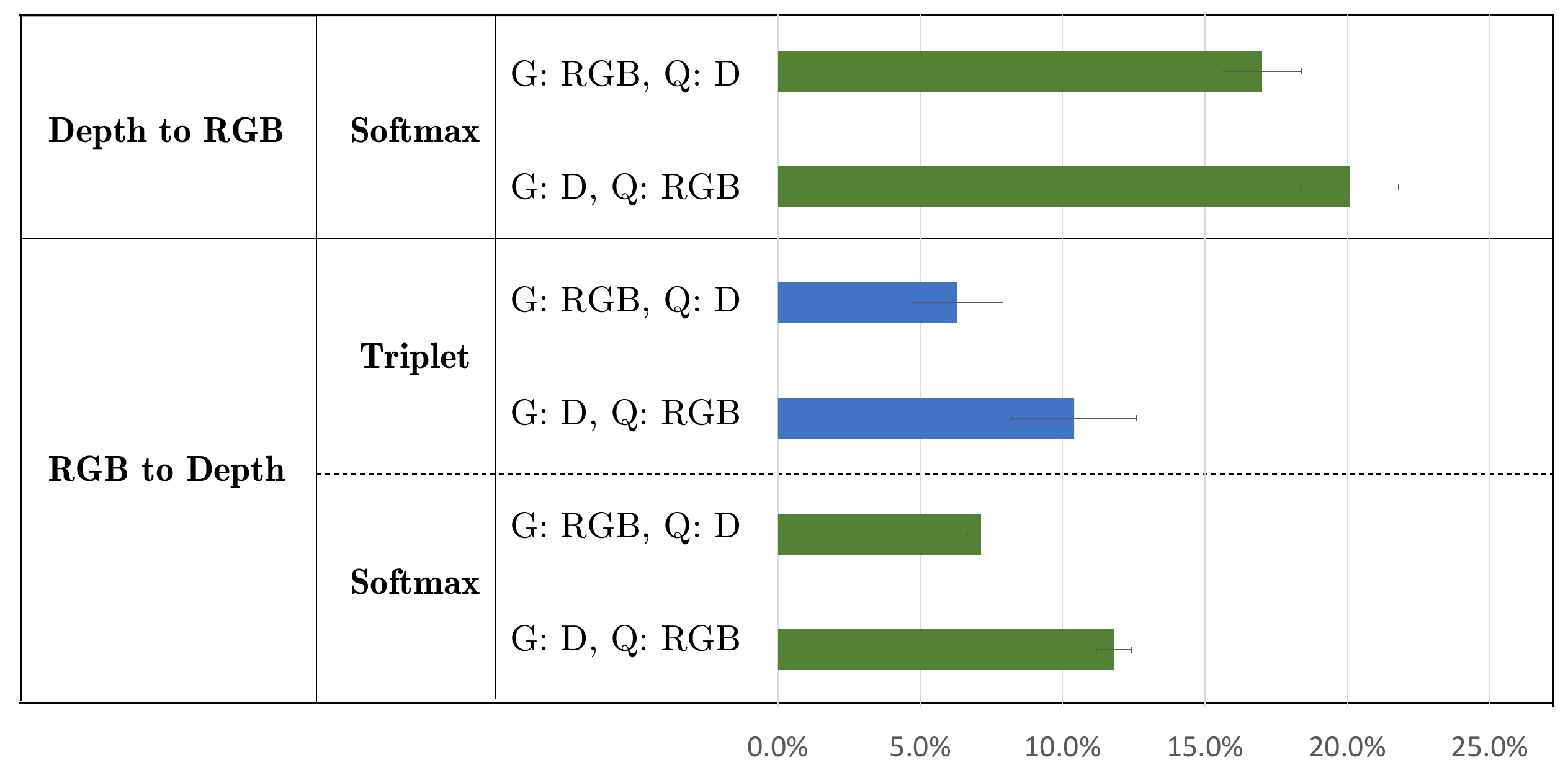}
\caption{Average mAP accuracy of various cross-modal distillation networks on the RobotPKU dataset. For all combinations we report varying query (Q) and gallery (G) modalities. The first column indicates the direction of the transfer for the cross-modal distillation. As no baseline for depth with triplet was successfully trained (see Table \ref{pkulabel}), no results reported.}
\label{robotpku_transresults}
\end{figure}

\begin{figure}[t!]
\centering
\includegraphics[width=0.9\linewidth]{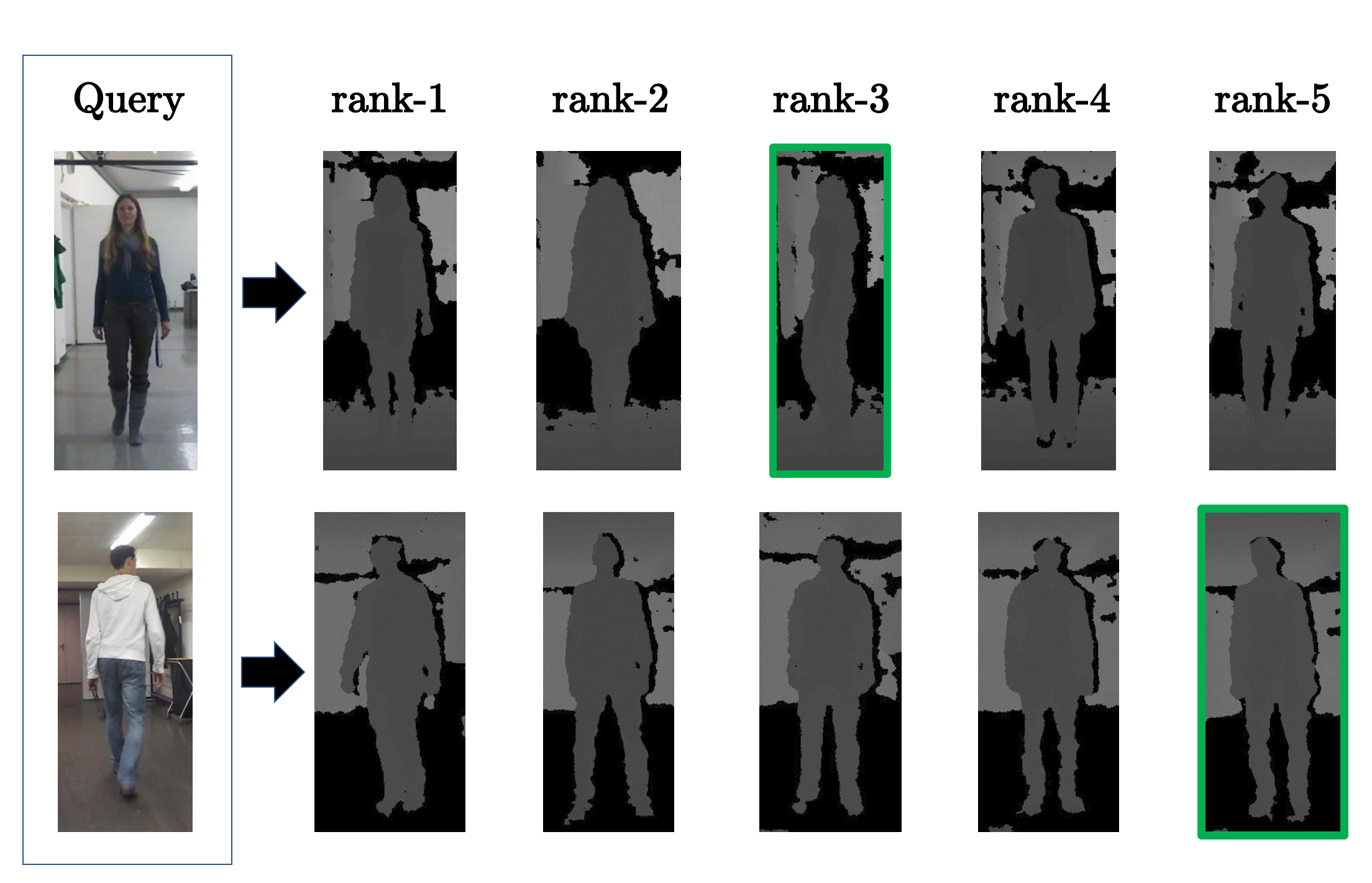}
\caption{Exemplary qualitative results for the proposed architecture on BIWI dataset. The green box denotes the correct match. Gallery (G) and Query (Q) varied for the modalities.}
\label{resultsbiwi}
\end{figure}

\begin{figure*}[t!]
\centering
\includegraphics[width=0.9\linewidth]{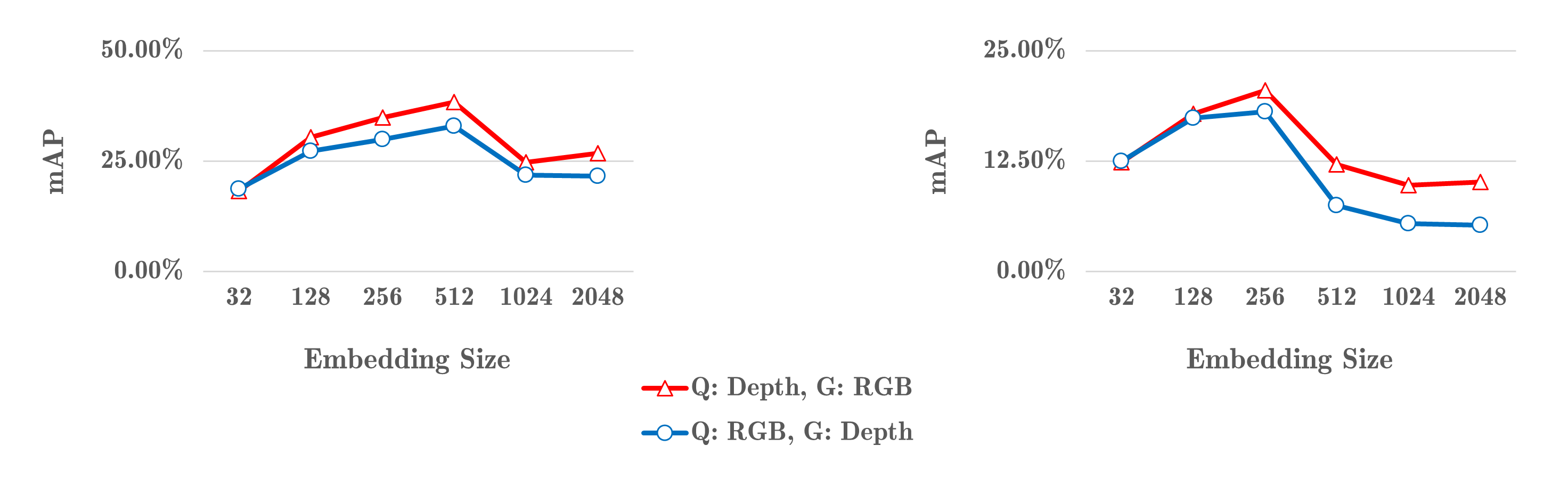}
\subfloat[Results for embedding size optimization for BIWI dataset.]{\hspace{.5\linewidth}}
\subfloat[Results for embedding size optimization for RobotPKU dataset.]{\hspace{.5\linewidth}}
\caption{Results for the optimization of the embedding size.}
\label{embsiz}
\end{figure*}

Figure \ref{biwi_transresults} and \ref{robotpku_transresults} present the average mAP accuracy of the networks trained with cross-modal distillation in the cross-modal tasks with varying population of query and gallery between RGB and depth trained on the BIWI dataset and RobotPKU dataset, respectively. The top two networks in the Figures train the baseline network in depth (step I.), and then transfer to RGB (step II.). The bottom networks train the baseline network in RGB (step I.), and then transfer to depth (step II.).

Results indicate that the accuracy obtained when transferring from RGB to depth are significantly lower than from depth to RGB. For the BIWI dataset using depth images to populate a reference gallery, and RGB images as query achieves a mAP accuracy of about 31\% using networks optimized with softmax loss. The best mAP accuracy for the same task and transferring from RGB to depth is about 13\%. Also for the RobotPKU dataset the transfer from depth to RGB significantly outperforms the transfer from RGB to depth. The difference of the best networks in mAP is 11\%/7.5\% for varying query and gallery population. An explanation for this behavior is that the general shape information of a person that is captured in depth data can, to a certain degree, be recovered in the RGB images. In contrast, the additional descriptive information which is inherent in RGB, like color information cannot be found in depth images. This behaviour will be further analyzed in section~\ref{sec:complex}.

The performance obtained for models trained with the two losses is only slightly differing (see Table~\ref{biwitab}). The overall best performance is obtained with a baseline in networks trained with softmax loss with an average mAP of 30.1\% with RGB as gallery (G) and depth (D) as query and 27.1\% for depth as gallery and RGB as query.    
Figure~\ref{resultsbiwi} shows an example of results for a network trained with cross-modal distillation on BIWI dataset, where the query image is RGB and the gallery image is depth. 
This Figure highlights the complexity of the task, as the differences between the gallery images are subtle.

In summary, to obtain the better results with cross-modal distillation, the transfer of knowledge should occur from depth to RGB. As shown in section \ref{sec:singmod} (Tables~\ref{biwitab} and~\ref{pkulabel}) in the single-modal task a much higher performance was obtained in the RGB modality. Hence, the performance in the single-modal task of the baseline network is not critical for the performance of cross-modal distillation. Our results suggest that the success of the distillation step is more dependent on the features learned from the modalities. Hence, the features learned in the depth modality were transferable to the RGB modality, while features learned in the RGB modality where not transferable to the depth modality. This gives an indication on the relation between the depth and RGB modality where the appearance information in depth is, to a certain degree, a subset of that found in RGB. Despite the indirect and direct nature of the loss functions, the results indicate that networks with a baseline trained with softmax loss and networks with a baseline network in triplet loss obtain similar results, which highlights the robustness of the method itself.

\begin{table*}[!ht]

\caption{Average accuracy of state-of-the-art and proposed networks for different scenarios on the BIWI dataset. Methods marked with (*) did not report all necessary details on their evaluation procedure, therefore their performance numbers are in brackets. \cite{zhuo2017} did not provide any details on their evaluation procedure. \cite{uddin2019} used different splits without a validation step (using 40 instead of 32 training instances as we do), thus their reported numbers might be too optimistic.
}
\resizebox{\linewidth}{!}
{\begin{tabular}{|l||l|l|l|l||l|l|l|l|}
\hline
{\textbf{Approach}}      & \multicolumn{4}{l||}{\textbf{Query-RGB, Gallery-Depth}}                     & \multicolumn{4}{l|}{\textbf{Query-Depth, Gallery-RGB}}                     \\ \cline{2-9} 
                                        & \textbf{rank-1 (\%)} & \textbf{rank-5 (\%)} & \textbf{rank-10 (\%)} & \textbf{mAP (\%)} & \textbf{rank-1 (\%)} & \textbf{rank-5 (\%)} & \textbf{rank-10 (\%)} & \textbf{mAP (\%)} \\ \hline \hline
\textbf{WHOS, Euclidean}\cite{Lisanti2015}      & 3.2 & 16.6 & 31.5 & 3.7    &  5.1 & 18.7 & 32.6 & 5.6                                   \\ \hline
\textbf{WHOS, XQDA}\cite{Lisanti2015}    &   8.4 & 31.7 & 50.2 & 7.9  & 11.6 & 34.1 & 51.4 & 12.1                         \\ \hline
\textbf{LOMO, Euclidean}\cite{liao2015person}                        & 2.8 & 16.4       & 32.5            &   4.8          & 3.3 & 15.6       & 29.8            &   5.6          \\ \hline
\textbf{LOMO, XQDA}\cite{liao2015person}                       & 13.7 & 43.2& 61.7            &   12.9          & 16.3 & 44.8      & 62.8            &   15.9         \\ \hline \hline
\textbf{Eigen-depth HOG/SLTP, CCA} \cite{zhuo2017}*                                    & (8.4)        & (26.3)       & (41.6)              &               - & (6.6)        & (27.6)       & (45.0)            & -              \\ \hline
\textbf{Eigen-depth HOG/SLTP, LSSCDL} \cite{zhuo2017}*                                 & (9.5)        & (27.1)       & (46.1)  &  -            & (7.4)        & (29.5)       & (50.3)       & -            \\ \hline
\textbf{Eigen-depth HOG/SLTP, Corr. Dict.} \cite{zhuo2017}*   
& (12.1)       & (28.4)       & (44.5)              &     -         & (11.3)       & (30.3)       & (48.2)             &  -            \\ \hline \hline
\begin{tabular}[c]{@{}l@{}}\textbf{Zero-padding network} \cite{wu2017} \end{tabular} &  5.9 $\pm$ 2.2  & 25.9 $\pm$ 6.4  & 47.1 $\pm$ 8.1  & 7.3 $\pm$ 4.0   &  10.3 $\pm$ 2.7  & 38.9 $\pm$ 6.5  & 62.8 $\pm$ 11.5  & 9.8 $\pm$ 3.8                         \\ \hline
\begin{tabular}[c]{@{}l@{}}\textbf{Two-stream network} \cite{wu2017} \end{tabular} &  9.0  $\pm$ 0.6   & 32.6  $\pm$ 1.1   & 55.3  $\pm$ 2.8   & 10.8  $\pm$ 0.8  & 7.9  $\pm$ 1.8   & 32.1  $\pm$ 2.4   & 54.1  $\pm$ 2.3   & 12.3  $\pm$ 1.6                 \\ \hline
\begin{tabular}[c]{@{}l@{}}\textbf{One-stream network} \cite{wu2017} \end{tabular} &  15.7 $\pm$ 0.8  & 50.3 $\pm$ 1.2  & 75.7 $\pm$ 0.5  & 16.9 $\pm$ 0.9     & 19.8 $\pm$ 0.3  & 55.7 $\pm$ 0.8  & 78.9 $\pm$ 1.1  & 23.8 $\pm$ 0.3                \\ \hline
\begin{tabular}[c]{@{}l@{}}\textbf{Local Shape} \cite{uddin2019}* \end{tabular} &  (36.5) & (79.7)  & (92.4)  & -    & (41.4)  & (82.5) & (94.4)  & --       \\ \hline

\begin{tabular}[c]{@{}l@{}}\textbf{Cross-modal distillation network,} \\ \textbf{Embedding size 128D, (ours) }\end{tabular}& \textbf{26.9 $\pm$ 1.8}  & \textbf{65.9 $\pm$ 2.3}  & \textbf{84.1 $\pm$ 3.1}  & \textbf{27.3 $\pm$ 1.7}             & \textbf{29.2 $\pm$ 2.3}  & \textbf{70.5 $\pm$ 2.3}  & \textbf{88.1 $\pm$ 0.9}  & \textbf{30.6 $\pm$ 2.0}    \\ \hline

\begin{tabular}[c]{@{}l@{}}\textbf{Cross-modal distillation network,} \\ \textbf{Embedding size 512D, (ours)}\end{tabular}& \textbf{32.6 $\pm$ 1.6}  & \textbf{70.9 $\pm$ 0.2}  & \textbf{88.8 $\pm$ 0.8}  & \textbf{33.0 $\pm$ 1.8}  & \textbf{36.6 $\pm$ 0.7}  & \textbf{76.7 $\pm$ 1.1}  & \textbf{92.3 $\pm$ 1.9}  & \textbf{38.4 $\pm$ 1.7}    \\ \hline

\begin{tabular}[c]{@{}l@{}}\textbf{Cross-modal distillation network + Attention,} \\ \textbf{Embedding size 512D, (ours)}\end{tabular}& \textbf{40.4 $\pm$ 2.1}  & \textbf{77.1 $\pm$ 1.7}  & \textbf{91.0 $\pm$ 1.0}  & \textbf{41.3 $\pm$ 1.8}  & \textbf{42.8 $\pm$ 3.9}  & \textbf{80.3 $\pm$ 1.6}  & \textbf{93.5 $\pm$ 1.0}  & \textbf{43.9 $\pm$ 3.9}    \\ \hline
\end{tabular}}
\label{sotabiwi}
\end{table*}

\begin{table*}[h!]
\caption{Average accuracy of state-of-the-art and proposed architecture for different scenarios on the RobotPKU dataset.}
\resizebox{\linewidth}{!}
{\begin{tabular}{|l||l|l|l|l||l|l|l|l|}
\hline
{\textbf{Approach}}      & \multicolumn{4}{l||}{\textbf{Query-RGB, Gallery-Depth}}                     & \multicolumn{4}{l|}{\textbf{Query-Depth, Gallery-RGB}}                     \\ \cline{2-9} 
                                        & \textbf{rank-1 (\%)} & \textbf{rank-5 (\%)} & \textbf{rank-10 (\%)} & \textbf{mAP (\%)} & \textbf{rank-1 (\%)} & \textbf{rank-5 (\%)} & \textbf{rank-10 (\%)} & \textbf{mAP (\%)} \\ \hline \hline
\textbf{WHOS, Euclidean}\cite{Lisanti2015}   &   3.8 & 16.3 & 29.5 & 3.9    & 3.5 &16.1 & 31.2& 5.4    \\ \hline
\textbf{WHOS, XQDA}\cite{Lisanti2015} & 10.0 & 31.8 & 49.8 & 8.2  &  9.8 & 31.0& 48.0 & 9.8 \\ \hline
\textbf{LOMO, Euclidean}\cite{liao2015person} &   3.6  & 15.0  & 28.0  & 3.9  &  3.7  & 15.3  & 28.7  & 4.9                           \\ \hline
\textbf{LOMO, XQDA}\cite{liao2015person} &   12.9  & 36.4  & 56.1  & 10.1  & 12.3  & 37.4  & 56.1  & 12.3                           \\ \hline \hline

\begin{tabular}[c]{@{}l@{}}\textbf{Zero-padding network}\cite{wu2017} \end{tabular} &   7.8 $\pm$ 0.9  & 29.0 $\pm$ 2.6  & 47.8 $\pm$ 3.3  & 7.7 $\pm$ 0.6   &  6.6 $\pm$ 0.6  & 26.8 $\pm$ 2.1  & 45.6 $\pm$ 2.8  & 8.3 $\pm$ 0.6                           \\ \hline
\begin{tabular}[c]{@{}l@{}}\textbf{Two-stream network}\cite{wu2017} \end{tabular} &  7.5  $\pm$ 1.2   & 29.1  $\pm$ 2.9   & 46.6  $\pm$ 2.7   & 7.7  $\pm$ 1.3  & 6.0  $\pm$ 2.0   & 24.6  $\pm$ 4.6   & 43.2  $\pm$ 5.1   & 8.7  $\pm$ 1.8                 \\ \hline
\begin{tabular}[c]{@{}l@{}}\textbf{One-stream network}\cite{wu2017}\end{tabular} &   11.9 $\pm$ 0.6  & 38.1 $\pm$ 1.0  & 57.3 $\pm$ 2.1  & 11.4 $\pm$ 0.5           &   12.5 $\pm$ 1.0  & 38.5 $\pm$ 1.5  & 56.7 $\pm$ 0.9  & 14.2 $\pm$ 1.4                    \\ \hline \hline

\begin{tabular}[c]{@{}l@{}}\textbf{Cross-modal distillation network,}\\ \textbf{Embedding size 128D, (ours)}\end{tabular}&  \textbf{17.5 $\pm$ 2.2}  & \textbf{51.9 $\pm$ 3.6}  & \textbf{72.7 $\pm$ 3.2}  & \textbf{17.1 $\pm$ 1.9}           & \textbf{19.5 $\pm$ 2.0}  & \textbf{54.3 $\pm$ 3.1}  & \textbf{74.4 $\pm$ 2.3}  & \textbf{19.8 $\pm$ 2.1}    \\ \hline 
\begin{tabular}[c]{@{}l@{}}\textbf{Cross-modal distillation network,}\\ \textbf{Embedding size 256D, (ours)}\end{tabular}&  \textbf{19.5 $\pm$ 1.0}  & \textbf{50.1 $\pm$ 0.5}  & \textbf{67.9 $\pm$ 0.7}  &\textbf{18.1 $\pm$ 1.2}          & \textbf{21.5 $\pm$ 1.1}  & \textbf{54.9 $\pm$ 1.4}  & \textbf{72.6 $\pm$ 1.0}  & \textbf{20.5 $\pm$ 1.0}    \\ \hline

\begin{tabular}[c]{@{}l@{}}\textbf{Cross-modal distillation network + Attention ,}\\ \textbf{Embedding size 256D, (ours)}\end{tabular}&  \textbf{25.3 $\pm$ 2.0}  & \textbf{58.1 $\pm$ 2.5}  & \textbf{74.7 $\pm$ 1.7}  &\textbf{23.5 $\pm$ 2.0}          & \textbf{22.9 $\pm$ 1.8}  & \textbf{56.0 $\pm$ 2.0}  & \textbf{72.8 $\pm$ 1.8}  & \textbf{22.4 $\pm$ 1.9} \\ \hline
\end{tabular}}
\label{sotapku}
\end{table*}

\subsection{Embedding Size Optimization}
\label{sec:embsize}
An important hyperparameter for the optimization of a re-identification system based on deep neural networks is the embedding size. In the preceding chapters an optimization with softmax loss from depth to RGB was shown to be most effective for both datasets. Hence, the ideal embedding size for this optimization is investigated. In Figure \ref{embsiz} the results of this analysis are shown. For  networks trained with deep distillation on the BIWI dataset the ideal embedding size is 512 which leads to a mAP of 33.04\%/38.44\%. For the RobotPKU dataset an ideal embedding size of 256 was found. Here, the best result in mAP is 18.13\%/20.52\%. The analysis underlines the relevant impact of the embedding size on the performance of the deep distillation method. The optimization of the embedding size explains the better performance of the distillation approach in comparison to \cite{hafner2019}. 

\subsection{Comparison with State-of-the-Art Methods}
\label{sec:compex}
In this section the results from section \ref{sec:crossdist} including the optimization in \ref{sec:embsize} are taken into a broader scope and are compared to existing methods for cross-modal person re-identification.
As deep learning based methods, one-stream network and zero-padding network as of \cite{wu2017} are analyzed. Additionally, a two-stream network similar to ~\cite{ye20181,ye20182} is evaluated. All deep neural network based methods are based on a Resnet50 architecture with an embedding size of 128 to fair comparison. The results of our proposed approach with embedding size 128 is reported in our previous work. Now, through the ablation study, the optimal embedding size for both datasets is obtained and reported in table 3 and 4, respectively, with and without cross-modal attention. As conventional approaches the WHOS feature extractor \cite{Lisanti2015} and the LOMO feature extractor \cite{liao2015person} will be investigated. The same features will be extracted for both modalities and are compared on the basis of Euclidean distance and the additional metric learning step Cross-view Quadratic Discriminant Analysis (XQDA). 
Additionally, the matching of Eigen-depth and HOG/SLTP features as reported by \cite{zhuo2017} and the local shape method \cite{uddin2019} is included in Table \ref{sotabiwi} for the BIWI dataset.

 Table~\ref{sotabiwi} presents the metrics of state-of-the-art and proposed networks for different scenarios on the BIWI dataset and in Table~\ref{sotapku} the results for the RobotPKU dataset are shown. First, it is apparent that the hand-crafted feature extractors lead to very low accuracy when matched in the Euclidean space. This is expected, as the modalities depth and RGB are heterogeneous and, hence, no direct comparison of hand-crafted features is possible. When applying the Cross-view Quadratic Discriminant Analysis (XQDA) the performance of the models based on hand-crafted features are significantly enhanced, while the LOMO features leads to the best results. For the BIWI dataset, these results also outperform the results from \cite{zhuo2017} for the Eigen-depth features combined with HOG/SILTP and also the results of deep learning based methods zero-padding network and two-stream network.
The deep learning based one-stream architecture is outperforming all methods based on hand-crafted features by at least 4\%/7\% for varying query and gallery in mAP accuracy for the BIWI dataset. Similarly, for RobotPKU it is superior by 1.3\%/1.9\%. Training with the cross-modal distillation idea enables an additional improvement compared to the one-stream network by 16.1\%/14.7\% for the BIWI dataset and 6.7\%/6.3\% for RobotPKU.

Overall results show that the network trained with the cross-modal distillation approach can significantly improve accuracy compared to state-of-the art methods for both BIWI and RobotPKU datasets.

\subsection{Experimental Analysis with Cross-modal Attention}
\label{sec:attention}
For this experimental analysis, we rely on the experimental findings of the previous section, namely that knowledge can transfer better from depth to RGB rather than from RGB to depth. Therefore, we use the RGB-stream as the backbone architecture, and the depth-stream to provide contextual information in the gating signal on the backbone architecture.
We run only the distillation step (step II.) for evaluating cross-modal attention.

Table~\ref{sotabiwi} and \ref{sotapku} report the experimental evaluation of cross-modal attention on the BIWI and RobotPKU datasets, respectively. Reported results clearly indicate the advantage of using cross-modal attention along with the distillation technique. For both datasets, integrating cross-modal attention improves the recognition accuracy by a significant margin in all  measures. Including cross-modal attention we outperform the reported accuracy of the local shape method \cite{uddin2019} in rank-1 accuracy. Since Uddin et al. \cite{uddin2019} do not report the used instance split and do not perform a validation (i.e., they use 40 instead of 32 training instances as we do), the improved rank-1 performance of our method indicates the superiority of our approach.

\subsection{Ablation Study}
The cross-modal distillation method is highly dependent on a successful knowledge transfer from depth to RGB. To get more insights into this transfer we evaluated the influence on network accuracy in the cross-modal tasks with varying components for knowledge transfer. Table \ref{influence} shows the impact of copying of weights, and freezing of mid to high-level layers on the accuracy. The ablation studies are done without cross-modal attention. Results are shown for the BIWI dataset with cross-modal distillation. If the freezing of mid- to high-level layers in the copied network is omitted, performance decreases by 10.7\%/2\% in mAP. Another reduction can be seen when the second network is not initialized with the weights of the first network. In this case the cross-modal performance in average mAP decreases by 18.0\%/9.7\% in comparison to the full transfer. These results underline the importance of each component for the training of a network with cross-modal distillation in performing knowledge transfer across the modalities.

\subsection{Qualitative Analysis of Neural Network Activations}
\label{sec:complex}

In this section an explanation for the superior performance of the network trained with distillation will be given, by analyzing deconvolution images of relevant deep learning methods, which are single-modal optimization as well as the two best-performing deep learning methods. 

\begin{figure}[t!]
\centering
\includegraphics[width=0.6\linewidth]{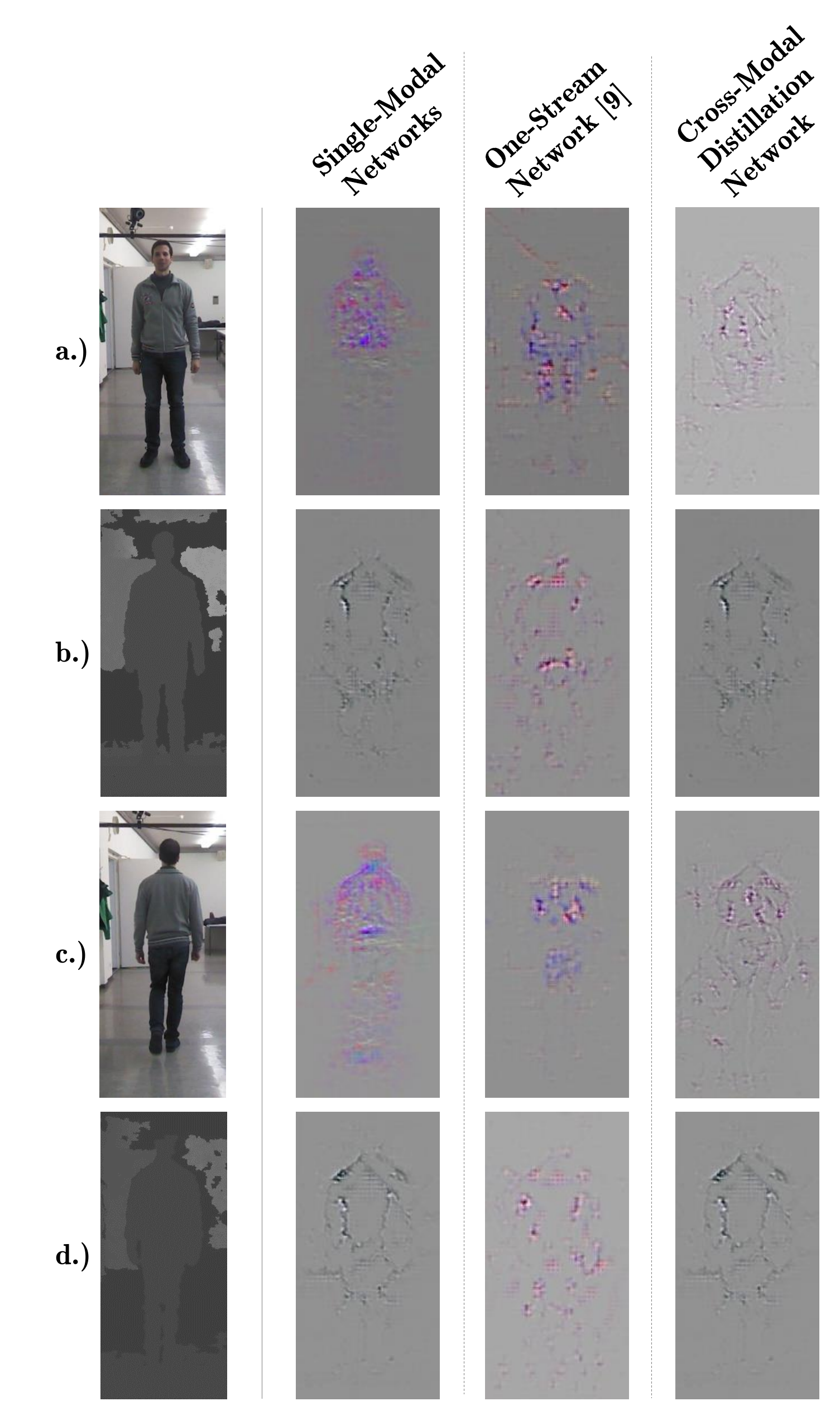}
\caption{Comparison of deconvolution images for different networks on BIWI data. Visualization is performed with guided backpropagation \cite{Springenberg2014}. Activation maps of cross-modal distillation in RGB highly differing to the other techniques.}
\label{deconv}
\end{figure}

Figure \ref{deconv} shows deconvolution images for different networks on two images from RGB (a. and c.) and depth (b. and d.) from the BIWI RGBD-ID dataset. The guided backpropagation algorithm was used for visualization of the activations for the networks \cite{Springenberg2014}. The shown architectures are trained separately for depth and RGB, the one-stream network, as the second in the state-of-the-art Table \cite{wu2017}, and our cross-modal distillation method.

\begin{table}[t!]
\centering
\caption{Ablation study of influence of various training scenarios for knowledge transfer. Results are average accuracy of the BIWI dataset for a network trained with cross-modal distillation and embedding size of 128.}
\resizebox{0.5\textwidth}{!}
{
\begin{tabular}{|l|l|l|l|l|l|}
\hline 
\multicolumn{2}{|l|}{\textbf{Scenario}}                                                                         & \textbf{rank-1 (\%)}          & \textbf{rank-5 (\%)}          & \textbf{rank-10 (\%)}         & \textbf{mAP(\%)}         \\ \hline \hline
\multirow{2}{*}{\textbf{\begin{tabular}[c]{@{}l@{}}No copying of weights,\\ No freezing of layers\end{tabular}}} & Q: RGB, G: D & 8.8 & 37.5 & 58.1 & 9.3      \\ \cline{2-6} 
                                                                                                                 & Q: D, G: RGB &   16.1 & 52.3 & 80.1 & 20.7             \\ \hline \hline
\multirow{2}{*}{\textbf{\begin{tabular}[c]{@{}l@{}}Copying of weights,\\ No freezing of layers\end{tabular}}}    & Q:RGB, G: D  & 16.7 & 49.8 & 69.6 & 16.6 \\ \cline{2-6} 
                                                                                                                 & Q: D, G: RGB & 25.9 & 68.7 & 89.4 & 28.5 \\ \hline \hline
\multirow{2}{*}{\textbf{\begin{tabular}[c]{@{}l@{}}Copying of weights,\\ Freezing of layers\end{tabular}}}       & Q:RGB, G: D  & 26.9  & 65.9  & 84.1  & 27.3 \\ \cline{2-6} 
                                                                                                                 & Q: D, G: RGB & 29.2  & 70.5  & 88.1  & 30.5 \\ \hline
\end{tabular}
}
\label{influence}
\end{table}

The images show that the activations for the different networks are varying considerably. When optimized for the single modalities, the networks in the RGB modality are activated by features inside the torso region of a person, like the color of the same. The network sensing in the depth modality is activated by the outer structure of the torso. For the one-stream network in the RGB modality the network is mostly activated by colors of torso and upper legs, while in the depth modality a cluttered outer structure of the torso is captured. For the RGB modality in the network trained with cross-modal distillation a very different activation map can be observed (images (a) and (c)). Instead of being activated by color features, we see that the network is mostly activated the structure of the torso for those images. Therefore, the knowledge from depth, which is a description of the problem with structural details, was transferred to the RGB modality. This finding shows that a transfer of knowledge between the modalities can benefit neural network training. As the describing features for the images are similar, the task of embedding to a common feature space is facilitated. This explains the better performance in cross-modal person re-identification as found in section 5.3.

\section{Conclusions}
\label{sec:conclusions}

In this paper, a new deep neural training scheme is proposed for cross-modal person re-identification that allow sensing between RGB and depth modalities. Its two-step approach enables the networks to exploit the relation between these two relevant modalities, and thereby provides a high level of performance.  
To further improve robustness of the feature representation, we propose to incorporate a cross-modal attention mechanism within the proposed distillation technique to dynamically select the more relevant convolutional filters  based on the privileged information from another modality, for robust feature representation and inference.

Experimental results on two public datasets indicate that our proposed network can outperform related state-of-the-art methods for cross-modal re-identification by up to 16.1\% in mAP. Results also show that features which are descriptive in the depth modality can successfully be extracted in the RGB modality for person re-identification. This implies that information captured in depth is to some extent retrievable in the RGB modality. 
Experimental analysis of the combination of the distillation technique and cross-modal gated attention shows significant performance improvement of 7.8$\%$ rank-01 and 12.2$\%$ mAP considering query- RGB and gallery-depth setting over the proposed distillation as well as all the state-of-the-art.

The analysis in this paper also showed that cross-modal person re-identification is a complex task, and the results in absolute numbers suggest that there is still room for improvement. In fact, the accuracies obtained in cross-modal re-identification (Tables \ref{sotabiwi} and \ref{sotapku}) are still lower than the accuracies for single-modal re-identification in the more difficult modality (Tables \ref{biwitab} and \ref{pkulabel}).

As one of the first works concerning RGB-depth re-identification, we have provided several initial insights into this problem, but still several open  questions and challenges remain.
It remains necessary to investigate the performance on data captured in real-world scenarios as opposed to controlled lab conditions as was done in this study.
Additionally, the absolute performance of any re-identification method is expected to decrease with an increasing number of instances in test sets and, hence, more possibilities for errors. 
For future work, it will therefore be necessary to obtain bigger and more diverse datasets for the RGB-depth use-case to facilitate data-hungry methods for robust feature extraction, and to develop new architectures that improve the embeddings for person matching across the different modalities \cite{sun2017}.
The publication of the SYSU-IR dataset in 2017 \cite{wu2017} pushed the interest in cross-modal person re-identification in RGB and infrared~\cite{ye20181,ye20182,dai2018}, and a similar effect could be expected for cross-modal person re-identification between RGB and depth.

\appendix

\label{sec:appenda}
\section{Split of Evaluation Datasets}

This appendix provides the label of individuals used to form the design (training set plus validation) and test subsets.

\subsection{BIWI RGBD-ID dataset:}  

Design set (Train + Validation set):\\[5pt]
			0,
            1,
            4,
            5,
            6,
            7,
            9,
            11,
            12,
            13,
            15,
            16,
            17,
            18,
            19,
            20,
            25,
            26,
            34,
            35,
            38,
            39,
            40,
            43,
            50,
            56,
            57,
            58,
            59,
            61,
            62,
            65,
            66,
            67,
            69,
            70,
            73,
            74,
            76,
            77.\\[5pt]
Test set:\\[5pt]
			2,
            3,
            8,
            10,
            14,
            21,
            22,
            23,
            24,
            27,
            28,
            29,
            30,
            31,
            32,
            33,
            36,
            37,
            41,
            42,
            44,
            45,
            46,
            47,
            48,
            49,
            51,
            52,
            53,
            54,
            55,
            60,
            63,
            64,
            68,
            71,
            72,
            75.

\subsection{RobotPKU dataset:}

Design set (Train + Validation set):\\[5pt]
0,
            2,
            3,
            15,
            16,
            18,
            19,
            20,
            21,
            22,
            23,
            25,
            27,
            28,
            29,
            30,
            31,
            32,
            33,
            34,
            35,
            36,
            37,
            41,
            43,
            44,
            45,
            46,
            47,
            52,
            54,
            55,
            58,
            59,
            60,
            63,
            66,
            67,
            68,
            72,
            73,
            74,
            77,
            78,
            80,
            82,
            83,
            84,
            87,
            88.\\[5pt]
Test set:\\[5pt] 
			1,
            4,
            5,
            6,
            7,
            8,
            9,
            10,
            11,
            12,
            13,
            14,
            17,
            24,
            26,
            38,
            39,
            40,
            42,
            48,
            49,
            50,
            51,
            53,
            56,
            57,
            61,
            62,
            64,
            65,
            69,
            70,
            71,
            75,
            76,
            79,
            81,
            85,
            86,
            89.

{\small

\bibliographystyle{plain}

}
\end{document}